\documentclass{bmvc2k}
\usepackage{multirow} 
\usepackage{enumitem}
\usepackage{caption}
\usepackage{amsfonts}

\title{HEAL: Learning-Free Source Free Unsupervised Domain Adaptation for Cross-Modality Medical Image Segmentation}
\addauthor{Yulong Shi}{yulongshi@stumail.neu.edu.cn}{1}
\addauthor{Jiapeng Li}{aprilt0621@gmail.com}{1}
\addauthor{Lin Qi}{qilin@bmie.neu.edu.cn}{1,2,3*}

\addinstitution{
College of Medicine and Biological Information Engineering, Northeastern University, Shenyang, China.
}
\addinstitution{
Key Laboratory of Medical Image Computing, Ministry of Education, Northeastern University, Shenyang, China.
}
\addinstitution{
Engineering Research Center of Medical Imaging and Intelligent Analysis, Ministry of Education, Northeastern University, Shenyang, China.
}

\runninghead{Shi et al.}{HEAL-SFUDA}

\begin{document}

\maketitle

\begin{abstract}
Growing demands for clinical data privacy and storage constraints have spurred advances in Source Free Unsupervised Domain Adaptation (SFUDA). SFUDA addresses the domain shift by adapting models from the source domain to the unseen target domain without accessing source data, even when target-domain labels are unavailable. However, SFUDA faces significant challenges: the absence of source domain data and label supervision in the target domain due to source free and unsupervised settings. To address these issues, we propose \textbf{HEAL}, a novel SFUDA framework that integrates \textbf{H}ierarchical denoising, \textbf{E}dge-guided selection, size-\textbf{A}ware fusion, and \textbf{L}earning-free characteristic. Large-scale cross-modality experiments demonstrate that our method outperforms existing SFUDA approaches, achieving state-of-the-art (SOTA) performance. The source code is publicly available at: \url{https://github.com/derekshiii/HEAL}.
\end{abstract}

\section{Introduction}
Medical image segmentation is crucial in diagnosis, treatment planning, and disease progression monitoring. Deep learning methods have shown remarkable success in this field \cite{havaei2017brain}\cite{isensee2021nnu}\cite{HE2025130732}. However, these methods typically require large labeled datasets for training, which are often difficult and expensive to acquire due to the need for expert annotation. Furthermore, data collected from different institutions using various imaging modalities or acquired from different devices can exhibit a significant domain shift, which significantly degrades the cross-modality generalization capability of models \cite{sandfort2019data}\cite{wachinger2018deepnat}.

In response to the increasing challenges of domain shift, unsupervised domain adaptation (UDA) has emerged as a crucial research direction. UDA strategically utilizes labeled data from the source domain to enable model adaptation in the completely unlabeled target domain \cite{oza2023unsupervised}. However, these methods require access to sensitive source domain data, which may raise privacy concerns or make collaboration between institutions difficult due to data sharing restrictions \cite{abadi2016deep}\cite{yang2019federated}. Therefore, source free unsupervised domain adaptation (SFUDA) has emerged as a promising paradigm in which a pre-trained source model is adapted to unlabeled target domains without accessing data from the source domain \cite{tian2024generation}. The source free and unsupervised settings are particularly appealing in medical image analysis, where data privacy is paramount. However, SFUDA presents significant challenges due to the inaccessibility of source domain data and the absence of guidance in target domain adaptation. Several SFUDA approaches focus on generating pseudo-labels in the target domain and using them for self-training \cite{dai2020curriculum,zou2018unsupervised}. Other approaches leverage generative models to generate source-like data for domain gap mitigation \cite{litrico2023guiding}, where diffusion models \cite{ho2020denoising} are promising in this paradigm \cite{zeng2024reliable}. 
However, existing SFUDA methods often involve self-training or fine-tuning the pre-trained model on the target domain, which can still raise concerns about computational cost and potential privacy leakage.

In this work, we propose a novel SFUDA framework (HEAL) for cross-modality medical image segmentation. \textbf{HEAL} incorporates \textbf{H}ierarchical denoising (HD), \textbf{E}dge-guided selection (EGS), size-\textbf{A}ware fusion (SAF), and \textbf{L}earning-free characteristic.

A key distinguishing feature of HEAL is the "learning-free" characteristic, which sets it apart from conventional SFUDA methods. In HEAL, the model is exclusively pre-trained on the source domain, and no further training, fine-tuning, or parameter updates are performed during domain adaptation to the target domain. Instead, the entire adaptation process is conducted purely through inference. This stands in stark contrast to many existing SFUDA approaches, which typically involve a target-domain learning phase, such as self-training with pseudo-labels\cite{dai2020curriculum,zou2018unsupervised}, iterative optimization to reduce domain shift, or training auxiliary networks to bridge domain gaps. By eliminating the need for such training, HEAL not only enhances computational efficiency and simplifies deployment but also preserves the integrity of the pre-trained source model and reduces the risks associated with learning from noisy or limited target data. The contributions of our study are summarized as follows: 
\begin{enumerate}[noitemsep]
    \item We introduce HEAL, a novel SFUDA framework for cross-modality medical image segmentation. HEAL can achieve effective domain adaptation without requiring further training or parameter updates to the pre-trained model, which inherently ensures superior data privacy and exceptional computational efficiency.
    \item We propose hierarchical denoising that refines pseudo-labels by sequentially leveraging entropy uncertainty and Normal-Inverse Gaussian (NIG) uncertainty, effectively mitigating error accumulation in the pseudo-labels.
    \item We introduce edge-guided selection to identify the most reliable samples generated by the diffusion model. By leveraging the inherent stochasticity of the diffusion model, multiple candidate images are synthesized, and the one with the highest structural consistency metric is selected as the most trustworthy.
    \item  We develop size-aware fusion that dynamically integrates pseudo-labels based on the size of the segmentation targets. This approach stabilizes the overall segmentation structure while enhancing the performance of small targets.
\end{enumerate}
\section{Method}
\begin{figure*}
\centering
\includegraphics[width=\textwidth]{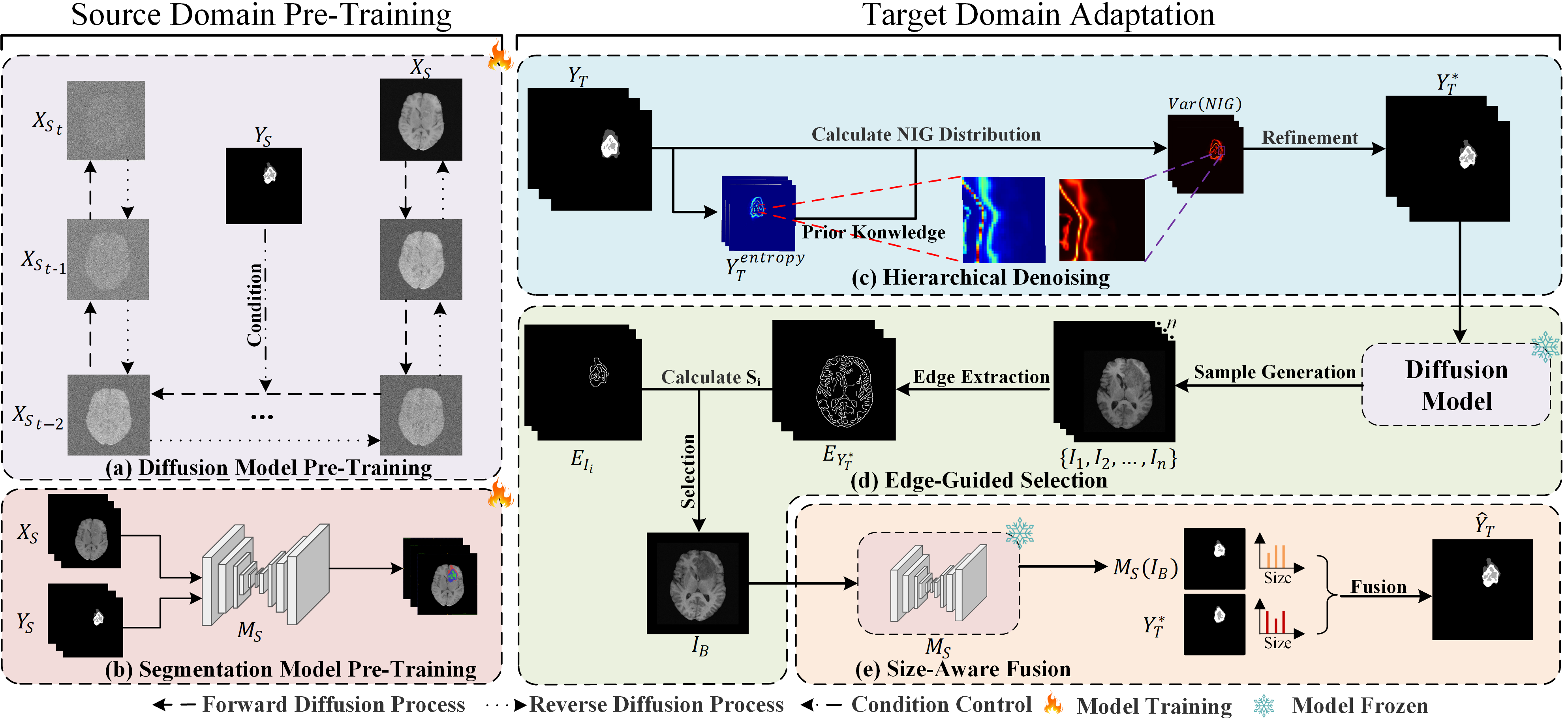}
\caption{\small An overview of HEAL, illustrating its key components: hierarchical denoising, edge-guided selection, and size-aware fusion.}

\label{fig1}
\end{figure*}
Figure~\ref{fig1} shows the overview architecture of HEAL. First, during the pre-training stage, we train a segmentation model $M_S$ and a diffusion model using source domain data $\{{X}_S,Y_S\}$, where ${X}_S$ are the source domain data and $Y_S$ are the label. Then, the segmentation model $M_S$ is employed to generate initial pseudo-labels $Y_T$ from the target domain data $X_T$. To mitigate error accumulation in pseudo-labels, we introduce HD to refine pseudo-labels utilizing entropy and NIG uncertainty maps. Subsequently, EGS is used to select the most reliable source-like sample $I_B$ by measuring the structural consistency. Finally, we design SAF to dynamically fuse the segmentations of EGS-selected samples $M_S(I_B)$ and the HD-refined pseudo labels $Y_T^*$.
\subsection{Hierarchical Denoising}
\subsubsection{Entropy Denoising:}
Given pseudo-labels $Y_T=M_S(X_T)$, we first compute voxel-wise entropy from the probability distribution of the initial pseudo-labels $Y_T$. Voxel-wise entropy is assigned to each class in each voxel as a measure of uncertainty, and higher entropy values indicate greater uncertainty. The voxel-wise entropy is used to guide the refinement of coarse error regions by filtering out voxels with high uncertainty. For a given voxel $v$ and a set of $C$ classes, let $P(v| c)$ denote the predicted probability of pseudo-label that voxel $v$ belongs to class $c$. The entropy $H(v)$ of the voxel $v$ is computed as:
\begin{equation}H(v)=-\sum_{c=1}^CP(c|v)\cdot\log_2\left(P(c|v)\right)\end{equation}
For each voxel $v$ in the target domain, the entropy-refined pseudo-labels $Y_T^{entropy}$ can be calculated:
\begin{equation}
Y_T^{entropy}(v) = Y_T(v) \cdot \mathbb{I}_{H(v) \leq \tau_1} =
\begin{cases}
    Y_T(v), & \text{if } H(v) \leq \tau_1 \\
    0,     & \text{if } H(v) > \tau_1
\end{cases}
\end{equation}
Where $\mathbb{I}_{\{{condition}\}}$ is the indicator function, and $\tau_1$ is the entropy threshold.
\subsubsection{Normal-Inverse Gaussian Variance Denoising:}
To further mitigate fine-grained pseudo-label errors, we introduce NIG denoising. Specifically, we model the class probability $p$ of each voxel in $Y_T$ as a Gaussian distribution $\mathcal{N}(\mu, \sigma^2)$, where the mean $\mu$ and variance $\sigma^2$ are governed by a NIG prior. This prior is derived from the entropy-refined pseudo-labels $Y_T^{entropy}$. The class probability $p$ can be calculated:
\begin{equation}p(\mu,\sigma^2\mid\alpha,\beta,\gamma,\omega)=\frac{\beta^{\alpha}\sqrt{\omega}}{\Gamma(\alpha)\sqrt{2\pi\sigma^{2}}} \left(\frac{1}{\sigma^{2}}\right)^{\alpha + 1} \exp\left\{-\frac{2\beta + \omega(\gamma - \mu)^{2}}{2\sigma^{2}}\right\}\end{equation}
Where $\Gamma(\cdot)$ is the gamma function and the NIG parameters $\gamma,\omega,\alpha,\beta$ are calculated from $Y_T^{entropy}$. Specifically, $\gamma$ is directly governed by $Y_T^{entropy}$, which anchors the distribution center to high-confidence regions. In contrast, $\alpha$ is modulated by the discrepancy between $Y_T$ and $Y_T^{entropy}$. Meanwhile, $\omega$ and $\beta$ are dynamically tuned based on the regional entropy $E(v)$, ensuring finer adjustments in regions of higher uncertainty. The NIG parameters can be calculated as follows:
\begin{equation}\alpha=\kappa\cdot\left|Y_T-Y_T^{entropy}\right|+\epsilon,\beta=\zeta_1(1-E(v))+\zeta_2,\omega=\frac{1}{\eta_1E(v)+\eta_2}\end{equation}
Where $\kappa, \zeta_1, \eta_1>0$ and $\epsilon, \zeta_2, \eta_2$ are small positive offsets. Here, we clarify that $E(v)$ is distinct from the voxel-wise entropy $H(v)$ used in the previous entropy denoising stage. While $H(v)$ quantifies uncertainty at the individual voxel level, $E(v)$ captures regional uncertainty by averaging entropy values within a local $3 \times 3$ voxel neighborhood around $v$. 

After we approximate the NIG distribution, the NIG uncertainty is quantified through the variance of NIG distribution to generate a binary mask for refining the $Y_T$. The HD-refined pseudo-label $Y_T^\ast$ can be obtained by:
\begin{equation}Var(\text{NIG})=\frac{\omega}{\beta(\alpha-1)}, Y_T^*(v)=Y_T(v)\cdot\mathbb{I}_{\left\{{Var(\text{NIG})}\leq\tau_2\right\}}\end{equation}
\subsection{Edge-Guided Selection}
Due to the inherent stochasticity of diffusion models, the quality of generated samples can vary significantly \cite{10.5555/3495724.3496767}. Therefore, selecting high-fidelity samples is crucial to ensure effective segmentation by the source pre-trained model \cite{ho2022classifierfreediffusionguidance}. To address this issue, we introduce EGS to select the most reliable generated sample based on the structural consistency metric $S_i$. First, we use the pre-trained diffusion model to generate $n$ samples conditioned on $Y_T^\ast$. Then, we apply $n$ sets of Canny edge detectors, where each set consists of two independent detectors: one for processing the generated sample $I_i$ (where $i = 1, 2, ..., n$) and the other for the corresponding diffusion model conditions $Y_T^\ast$:
\begin{equation}\{\{E_{I_{1}},E_{Y_{T}^{*}}\},\{E_{I_{2}},E_{Y_{T}^{*}}\},...,\{E_{I_{n}},E_{Y_{T}^{*}}\}=\mathrm{Canny}(\{I_{1},I_{2},...,I_{n}\},Y_{T}^{*})\end{equation}
Where $E_{I_i}$ represents the edge map of the $i_{th}$ generated image $I_i$. $E_{Y_T^\ast}$ represents the edge map of the corresponding diffusion model conditions $Y_T^\ast$. $\mathrm{Canny}(\cdot,\cdot)$ denotes the application of the Canny edge detector. To quantify the consistency between the generated image and the conditions, we calculate the structural consistency metric $S_i$ for each generated image $I_i$:
\begin{equation}S_{i}=\frac{|E_{I_{i}}\cap E_{Y_{T}^{*}}|}{|E_{Y_{T}^{*}}|}\end{equation}
Where $S_i$ is the structural consistency metric for the $i_{th}$ generated image $I_i$. $|E_{I_i}\cap E_{Y_T^\ast}|$ represents the number of aligned edge pixels between the generated image $I_i$ and the diffusion condition $Y_T^\ast$. $|E_{Y_T^\ast}|$ represents the total number of edge pixels in the diffusion model conditions $Y_T^\ast$. 
A higher $S_i$ indicates greater consistency between the generated sample and the conditions, suggesting a more reliable sample. After calculating the structural consistency $S_i$, we select $I_B$ with the highest structural consistency $S_i$ for subsequent segmentation.
\subsection{Size-Aware Fusion}
The EGS-selected sample $I_B$ demonstrates superior structural preservation for large anatomical structures due to its global consistency modeling \cite{10.5555/3600270.3602196}, whereas the HD-refined pseudo-labels $Y_T^\ast$ provide reliable guidance for small objects \cite{xu2024deep}. Therefore, we propose SAF that dynamically selects the most reliable source for each target according to the size. Specifically, we fuse the categories with the smallest proportion of voxels in $Y_T^\ast$ with other categories in $M_S(I_B)$. The final segmentation result $\widehat{Y}_T$ can be obtained by:
\begin{equation}\widehat{Y}_T=\sum_{k=1}^C\lambda_k\cdot[\mathbb{I}(v_k=v_{min})\cdot Y_T^*+\mathbb{I}(v_k\neq v_{min})\cdot M_S(I_B)]\end{equation}
Where $v_{min}=min\{v_1,v_2,...,v_{C}\}$ denotes the fewest voxels of the target category, $M_S(I_B)$ is the segmentation of the EGS-selected sample $I_B$ , and $\lambda_k=\frac{v_k^{-1}}{\Sigma v_i^{-1}}$ is the dynamic weight of the smallest target.
\section{Experiments and Results}
\subsection{Experimental Setup}
\subsubsection{Dataset and Evaluation Metrics:}
We validate our method on the BraTS 2021 dataset \cite{menze2014multimodal}, Kvasir-SEG \cite{kvasir}, and CVC-ClinicDB \cite{CVCDB}. \textbf{BraTS 2021} provides 1,251 fully annotated 3D MRI volumes with four modalities: T1, contrast-enhanced T1 (T1ce), T2, and FLAIR. The segmentation targets within this dataset encompass three tumor subregions: whole tumor (WT), tumor core (TC), and enhancing tumor (ET). \textbf{Kvasir-SEG} is a widely used benchmark dataset consisting of 1,000 endoscopic images with pixel-wise annotated segmentation masks. \textbf{CVC-ClinicDB} comprises 612 still frames extracted from multiple colonoscopy video sequences with corresponding binary masks delineating the polyp regions.

We evaluate HEAL on two medical image segmentation tasks across four domain adaptation directions: T1$\rightarrow$T1ce and T2$\rightarrow$FLAIR for brain tumor segmentation, and Kvasir-SEG$\rightarrow$CVC-ClinicDB and CVC-ClinicDB$\rightarrow$Kvasir-SEG for polyp segmentation. Segmentation performance is quantitatively assessed using two standard metrics: the Dice similarity coefficient (Dice) \cite{dice} and the average surface distance (ASD) \cite{asd}. For the polyp segmentation tasks, due to their single-class nature, HEAL is applied without the size-aware fusion, employing only hierarchical denoising and edge-guided selection. As the target domain is not used during training, we do not partition it into training and testing subsets. Instead, the model is directly performing inference on the full target domain to assess its adaptation performance.

\subsubsection{Implementation Details:}All experiments were implemented using the PyTorch framework and conducted on an NVIDIA RTX 4090 GPU with 24GB memory. We used Med-DDPM \cite{10493074} with a noise schedule of 250 time steps $t$, and the nnUNet framework \cite{isensee2021nnu}, which employs a ResNet-based U-Net architecture, trained for 300 epochs. Both the entropy threshold $\tau_1$ and the NIG distribution variance threshold $\tau_2$ were set to 0.2. In EGS, the number of generated samples $n$ is 6, and the threshold of the Canny edge detector is 0.1.
\subsection{Comparison with SOTA Methods}
In our experiments, \textbf{No Adaptation} indicates training on the source domain and directly testing on the target domain without adaptation. \textbf{Supervised} indicates training and testing on the target domain. Furthermore, we compared HEAL with four SOTA SFUDA methods: 1) ProtoContra \cite{protocontra}, 2) DPL \cite{DPL}, 3) UPL \cite{upl}, and 4) IAPC \cite{IAPC}. We reproduce all the comparison methods according to their official codes and configurations.
\begin{table}[t!] 
\centering
\footnotesize 
{
\setlength{\tabcolsep}{2.7pt}
\begin{tabular}{|l|cccc|cccc|cccc|cccc|}
\hline
\multirow{3}{*}{Method} & \multicolumn{8}{c|}{T1$\rightarrow$T1ce} & \multicolumn{8}{c|}{T2$\rightarrow$FLAIR} \\
\cline{2-17}
& \multicolumn{4}{c|}{Dice (\%) $\uparrow$} & \multicolumn{4}{c|}{ASD (mm) $\downarrow$} & \multicolumn{4}{c|}{Dice (\%) $\uparrow$} & \multicolumn{4}{c|}{ASD (mm) $\downarrow$} \\
\cline{2-5}\cline{6-9}\cline{10-13}\cline{14-17}
& WT & TC & ET & Mean & WT & TC & ET & Mean & WT & TC & ET & Mean & WT & TC & ET & Mean \\
\hline
No Adaptation & 64.3 & 57.2 & 55.1 & 58.9 & 3.3 & 2.0 & 3.2 & 2.8 & 67.2 & 63.0 & 50.9 & 60.4 & 3.2 & 5.1 & 2.4 & 3.6 \\
Supervised    & 86.2 & 93.1 & 73.2 & 84.2 & 1.3 & 0.6 & 1.4 & 1.1 & 93.0 & 84.3 & 81.1 & 86.1 & 0.8 & 1.4 & 1.0 & 1.1 \\
\hline
ProtoContra\cite{protocontra} & 46.1 & 47.9 & 33.4 & 42.5 & 6.0 & 5.5 & 4.0 & 5.2 & 63.3 & \underline{56.7} & \underline{37.6} & \underline{52.5}& 4.3 & 5.8 & \underline{4.4} & \underline{4.8} \\
DPL\cite{DPL} & 54.3 & 49.9 & 44.8 & 49.7 & 5.3 & 3.7 & 3.6 & 4.2 & 58.8 & 36.4 & 18.5 & 37.9 & \underline{4.0} & 7.0 & 7.8 & 6.3 \\
IAPC\cite{IAPC} & 59.4 & 58.8 & 34.8 & 51.0 & 7.3 & 6.1 & 6.1 & 6.5 & \underline{65.1} & 53.4 & 30.9 & 49.8 & 4.5 & \underline{5.4} & 5.8 & 5.2 \\
UPL\cite{upl} & \underline{66.1} & \underline{67.7} & \underline{60.0} & \underline{64.6} & \underline{4.9} & \underline{3.6} & \underline{2.8} & \underline{3.7} & 54.0 & 48.4 & 27.7 & 43.4& 9.9 & 7.4 & 6.5 & 7.9  \\
\textbf{HEAL} & \textbf{80.7} &{\textbf{82.9}} &{\textbf{68.1}} & {\textbf{77.3}} &{\textbf{2.5}} &{\textbf{1.5}} &{\textbf{2.0}} & {\textbf{2.0}} &{\textbf{82.9}} & {\textbf{74.5}} & {\textbf{63.0}} & {\textbf{73.5}} & {\textbf{2.8}} & {\textbf{2.6}} & {\textbf{2.4}} & {\textbf{2.6}} \\
\hline
\end{tabular}
}
\caption{\small Performance comparison of SFUDA methods for brain tumor segmentation. Best results are in \textbf{bold}, second best are \underline{underlined}.} 
\label{table-briantumor}
\end{table}

\begin{table}[t!] 
\centering 
\footnotesize 
{ 
\setlength{\tabcolsep}{5pt} 
\begin{tabular}{|l|c|c|c|c|} 
\hline
\multirow{2}{*}{Method} & \multicolumn{2}{c|}{Kvasir-SEG$\rightarrow$CVC-ClinicDB} & \multicolumn{2}{c|}{CVC-ClinicDB$\rightarrow$Kvasir-SEG} \\
\cline{2-5} 
& Dice (\%) $\uparrow$ & ASD (mm) $\downarrow$ & Dice (\%) $\uparrow$ & ASD (mm) $\downarrow$ \\
\hline
No Adaptation & 65.1$\pm$21.5 & 5.8$\pm$11.1 & 60.2$\pm$27.0 & 23.7$\pm$20.4  \\
Supervised    & 99.8$\pm$0.1 & 0.36$\pm$0.1 & 97.4$\pm$1.3 & 0.87$\pm$1.1  \\
\hline
ProtoContra\cite{protocontra} & \underline{78.0$\pm$12.1} & \textbf{0.9$\pm$0.6} & \textbf{68.9$\pm$26.4} & 9.5$\pm$15.0 \\
DPL\cite{DPL} & 67.2$\pm$24.0 & 8.0$\pm$12.4 & 66.1$\pm$33.1 & \textbf{7.8$\pm$17.7} \\
IAPC\cite{IAPC} & 73.3$\pm$29.6 & 4.4$\pm$7.9 & 61.2$\pm$26.3 & 13.7$\pm$13.2 \\
UPL\cite{upl} & 67.1$\pm$14.6 & \underline{2.2$\pm$0.9} & 60.2$\pm$31.3 & 11.4$\pm$17.2 \\
\textbf{HEAL} & \textbf{81.8$\pm$22.6} & 4.3$\pm$8.2 & \underline{66.5$\pm$28.8} & \underline{9.15$\pm$11.6} \\
\hline
\end{tabular}
} 
\caption{\small Performance comparison of SFUDA methods for polyp segmentation. Best results are in \textbf{bold}, second best are \underline{underlined}.} 
\label{table-polyp}
\end{table}
The quantitative evaluation results for cross-modality brain tumor segmentation are presented in Table \ref{table-briantumor}. A substantial performance gap is observed between the \textbf{Supervised} and \textbf{No Adaptation} baselines in both directions, highlighting the severe domain shifts between the source and target domains. In both directions, \textbf{HEAL} remarkably outperforms all other SFUDA approaches on WT, TC, and ET, achieving the highest mean Dice of 77.3\%, 73.5\%, and the lowest mean ASD of 2.0mm, 2.6mm in T1$\rightarrow$T1ce and T2$\rightarrow$FLAIR, respectively. Moreover, compared to \textbf{No Adaptation}, the competitive performance of \textbf{HEAL} on both Dice and ASD can be attributed to the synergistic effect of HD, EGS, and SAF. 

Table~\ref{table-polyp} presents the quantitative results for cross-modality polyp segmentation. There is a clear performance gap between the \textbf{Supervised}, with Dice scores of 99.8\% and 97.4\%, and the \textbf{No Adaptation}, which achieves Dice scores of only 65.1\% and 60.2\% in the Kvasir-SEG$\rightarrow$CVC-ClinicDB and CVC-ClinicDB$\rightarrow$Kvasir-SEG directions, respectively. This substantial gap highlights the challenges posed by domain shift, thereby underscoring the necessity for effective adaptation methods such as HEAL, which demonstrates strong domain adaptation capability. Specifically, in the Kvasir-SEG$\rightarrow$CVC-ClinicDB direction, \textbf{HEAL} achieves the best Dice of 81.8\%, a remarkable improvement over the \textbf{No Adaptation} and surpassing other SFUDA methods. In the reverse direction, CVC-ClinicDB$\rightarrow$Kvasir-SEG, \textbf{HEAL} also delivers highly competitive performance, obtaining a Dice of 66.5\% and an ASD of 9.15mm, which are marked as second-best results among the compared adaptation techniques. These results demonstrate \textbf{HEAL}'s effectiveness in significantly mitigating domain shift and enhancing segmentation accuracy across different polyp datasets and modalities without relying on source data. 
\begin{figure*}[!t]
\centering
\includegraphics[width=\textwidth]{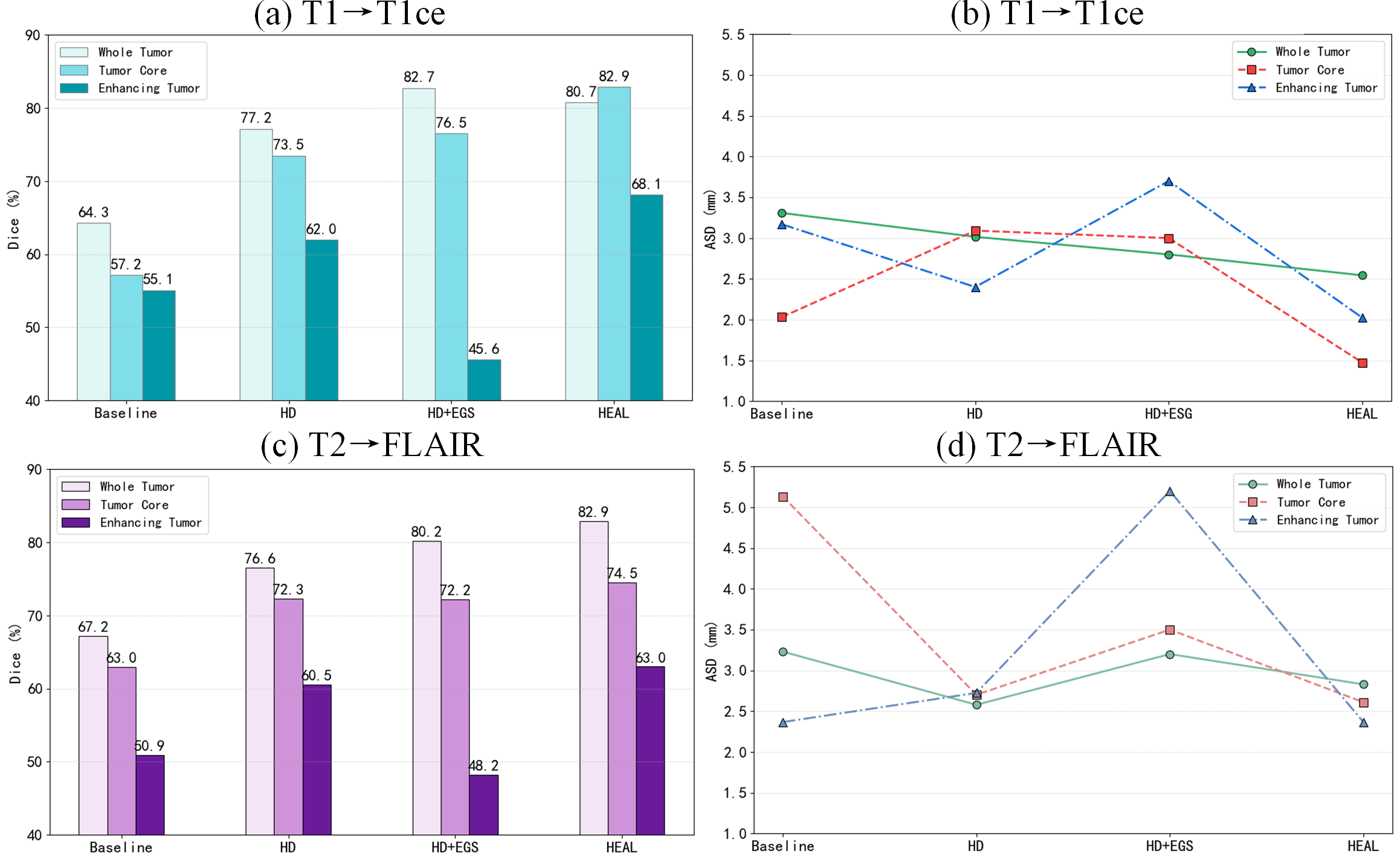}
\caption{\small Ablation study of each component in HEAL on brain tumor segmentation. HD refers to the results obtained by hierarchical denoising, HD+EGS denotes the combination of hierarchical denoising and edge-guided selection, and HEAL represents the full method with hierarchical denoising, edge-guided selection, and size-aware fusion.}
\label{fig2}
\end{figure*}
\subsection{Ablation Study}
\begin{table}[t!] 
\centering 
\footnotesize 
{ 
\setlength{\tabcolsep}{5pt} 
\begin{tabular}{|l|c|c|c|c|} 
\hline
\multirow{2}{*}{Method} & \multicolumn{2}{c|}{Kvasir-SEG$\rightarrow$CVC-ClinicDB} & \multicolumn{2}{c|}{CVC-ClinicDB$\rightarrow$Kvasir-SEG} \\
\cline{2-5} 
& Dice (\%) $\uparrow$ & ASD (mm) $\downarrow$ & Dice (\%) $\uparrow$ & ASD (mm) $\downarrow$ \\
\hline
Baseline & 65.1$\pm$21.5 & 5.8$\pm$11.1 & 60.2$\pm$27.0 & 23.7$\pm$20.4  \\
Entropy-refined & 72.6$\pm$24.5 & 5.8$\pm$12.2 & 63.1$\pm$33.0 &  17.9$\pm$39.6\\
HD & \underline{76.4$\pm$21.3} & \underline{5.1$\pm$10.9} & \underline{63.8$\pm$35.4} & \underline{16.2$\pm$20.8} \\
\textbf{HEAL} (HD+EGS) & \textbf{81.8$\pm$22.6} & \textbf{4.3$\pm$8.2} & \textbf{66.5$\pm$28.8} & \textbf{9.15$\pm$11.6} \\
\hline
\end{tabular}
} 
\caption{\small Ablation study of HEAL on polyp segmentation. Entropy-refined refers to the pseudo-labels refined by voxel-wise entropy. Best results are in \textbf{bold}, second best are \underline{underlined}. }

\label{table-ablation}
\end{table}
\begin{figure*}[!t]
\centering
\includegraphics[width=\textwidth]{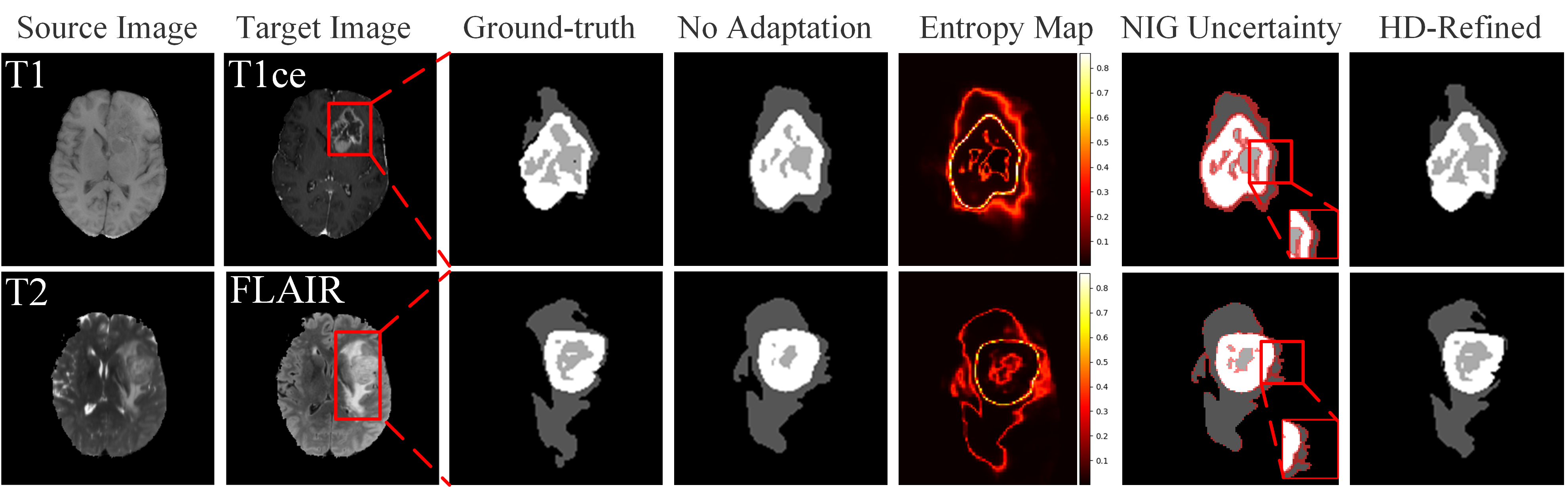}
\caption{\small Qualitative ablation study of hierarchical denoising on brain tumor segmentation. The first row corresponds to patient \#0000, and the second row to patient \#0051.}
\label{fig3}
\end{figure*}
To assess the individual contributions of each component in HEAL, we perform ablation studies on brain tumor segmentation and polyp segmentation. For brain tumor segmentation, we evaluate four configurations, \textbf{Baseline} (No Adaptation), \textbf{HD}, \textbf{HD+EGS}, and \textbf{HEAL} (HD+EGS+SAF), under two adaptation adaptations: T1$\rightarrow$T1ce and T2$\rightarrow$FLAIR. For polyp segmentation, we conduct ablation under two adaptation directions: Kvasir-SEG$\rightarrow$CVC-ClinicDB and CVC-ClinicDB$\rightarrow$Kvasir-SEG, using four configurations: \textbf{Baseline}, \textbf{Entropy-refined}, \textbf{HD}, and \textbf{HEAL} (HD+EGS).

As shown in Figure~\ref{fig2} (a), in the T1$\rightarrow$T1ce direction, \textbf{HEAL} demonstrably enhances segmentation performance, yielding significant Dice improvements over the \textbf{Baseline} of 16.4\%, 25.7\%, and 13.0\% for WT, TC, and ET, respectively. Figure~\ref{fig2} (b) reveals that \textbf{HD} reduces ASD for WT relative to the \textbf{Baseline}. Furthermore, the integration of EGS and SAF in \textbf{HEAL} further diminishes ASD to 2.8mm and 2.6mm, respectively. Similarly, Figure~\ref{fig2} (c) demonstrates that \textbf{HEAL} achieves Dice gains of 15.7\%, 11.5\%, and 12.1\% compared to the \textbf{Baseline} in the FLAIR$\rightarrow$T2 direction. As visualized in Figure~\ref{fig2} (d), \textbf{HEAL} effectively minimizes ASD across all tumor sub-regions in the FLAIR$\rightarrow$T2 direction. These results collectively underscore the significant and consistent improvements in segmentation performance achieved by \textbf{HEAL} across different modalities.

Table~\ref{table-ablation} demonstrates that \textbf{HEAL} consistently enhances polyp segmentation performance across both adaptation directions. Beginning with the \textbf{Baseline}, the \textbf{Entropy-refined} led to a substantial Dice improvement of 7.5\%, particularly in the Kvasir-SEG$\rightarrow$CVC-ClinicDB direction. As the entropy-refined pseudo-labels provide reliable prior knowledge for subsequent NIG denoising, \textbf{HD} further enhances segmentation performance, leading to consistent improvements in Dice and reductions in ASD across both directions. Finally, integrating \textbf{EGS} with \textbf{HD}, \textbf{HEAL} achieves the best overall results, with Dice improvements of 16.7\% and 6.3\% over the baseline in Kvasir-SEG$\rightarrow$CVC-ClinicDB and CVC-ClinicDB$\rightarrow$Kvasir-SEG, respectively. Additionally, ASD is substantially reduced, particularly in the CVC-ClinicDB$\rightarrow$Kvasir-SEG. These results highlight the effectiveness and complementary contributions of hierarchical denoising and edge-guided selection within \textbf{HEAL}.

Figure \ref{fig3} visually demonstrates the effectiveness of HD in the refinement of pseudo-labels. In the \textbf{No Adaptation} column, we observe that the segmentation results exhibit noticeable inaccuracies compared to the \textbf{Ground-truth}. The \textbf{Entropy Map} is a heatmap that highlights the regions of entropy uncertainty, primarily concentrated around the coarse errors. Subsequently, the \textbf{NIG Uncertainty} further refines these uncertainty regions, pinpointing areas requiring more precise error correction, as highlighted by the red bounding boxes. This qualitative assessment underscores the critical role of HD in enhancing the reliability of pseudo-labels and improving the overall segmentation performance of HEAL.

\begin{figure*}[t]
\centering
\includegraphics[width=\textwidth]{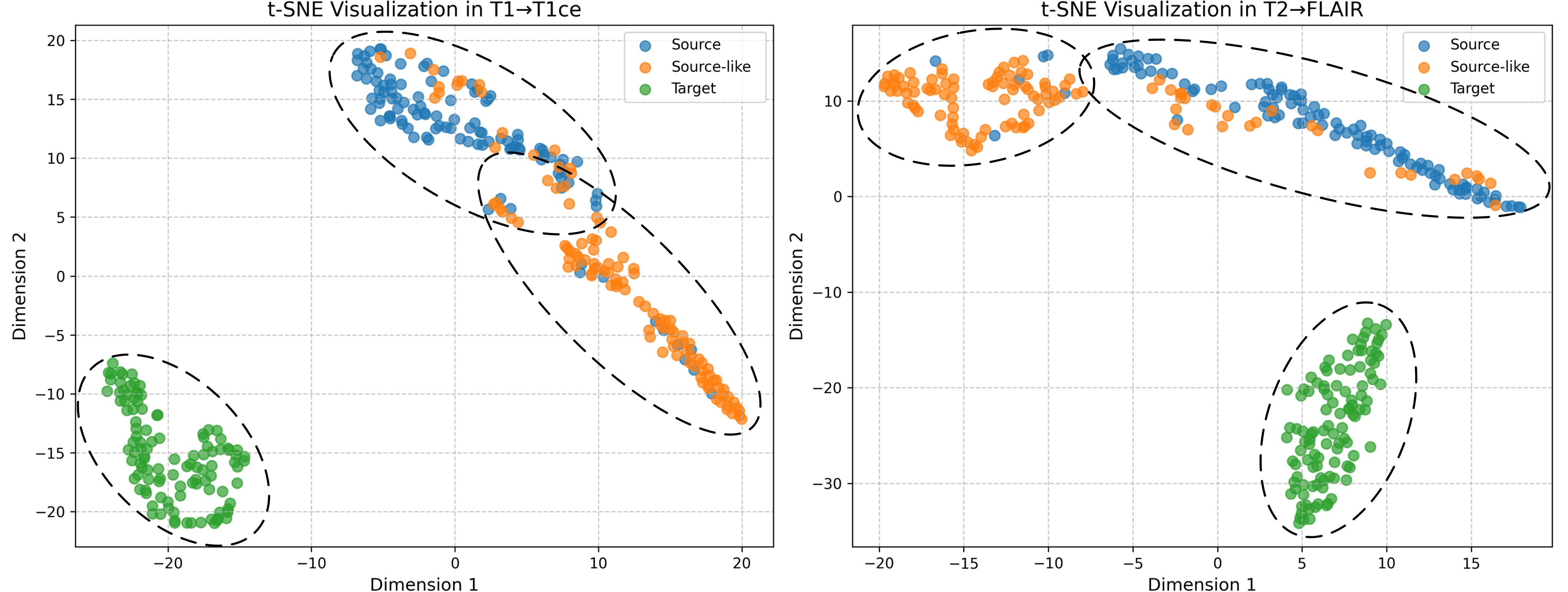}
   \caption{\small t-SNE visualizations of feature distribution in T1$\rightarrow$T1ce and T2$\rightarrow$FLAIR on BraTS2021.}
\label{fig_tsne}
\end{figure*}

To further investigate the effectiveness of HEAL in aligning feature distributions across domains, we visualize the feature distribution using t-SNE, as shown in Figure \ref{fig_tsne}. In both adaptation directions, the feature representations of samples from the source domain, the target domain, and source-like samples generated by the diffusion model are projected into a 2D space. The visualizations demonstrate that the source-like features lie closer to the source domain than the target domain, indicating that the diffusion model effectively re-renders the target domain's structural information into a new representation that is stylistically aligned with the source domain. By leveraging the source knowledge captured during diffusion model pretraining, HEAL mitigates the risks of over-segmentation and information loss that may arise from hierarchical denoising, ultimately facilitating more robust and accurate domain adaptation.

\section{Conclusion}
In this work, we proposed HEAL, a novel SFUDA framework for cross-modality medical image segmentation without any training when performing target domain adaptation. The hierarchical denoising effectively mitigates error accumulation through entropy and NIG denoising, then the edge-guided selection ensures diffusion-generated samples preserve the critical anatomical structure and transfer the target domain knowledge to the source domain. Finally, the size-aware fusion effectively fuses the reliable classes in the HD-refined pseudo-labels and the segmentation of the EGS-selected sample. Experimental results on brain tumor and polyp segmentation tasks validate the superior performance of our method compared to existing SFUDA approaches.

A key aspect of HEAL's mechanism is that the framework does not attempt to make the source model directly interpret the target image. Instead, it leverages the structural information encapsulated in the HD-refined pseudo-label. The diffusion model, conditioned on the target image, is tasked with generating a source-like sample that is stylistically consistent with the source domain but structurally aligned with the target's anatomy. Therefore, the segmentation model is not processing the target image it has never seen; it is segmenting a synthesized source-like image where the tumor structures have been rendered according to the guidance from. This process effectively translates the adaptation challenge from a difficult cross-modality recognition problem into a more manageable in-domain segmentation task for the frozen source model.

\noindent \textbf{limitations}:
The effectiveness of HEAL is inherently linked to the generalization capability of the pre-trained segmentation model. When the domain shift is excessively large, the initial pseudo-labels may be of low quality, which propagate errors through the subsequent process and ultimately degrade HEAL’s performance. Furthermore, the learning-free nature of our approach may limit its ability to capture nuanced or novel target-specific features that could otherwise be addressed by methods involving target-domain parameter update.
\section{Acknowledgments}
This work is supported by the Key Research and Development Program of Liaoning Province (2024JH2/102500076).
\bibliography{egbib}
\end{document}